\documentclass[letterpaper]{article}
\usepackage{aaai19}
\usepackage{times}
\usepackage{helvet}
\usepackage{courier}
\usepackage{url}
\usepackage{xcolor}
\usepackage{graphicx}
\usepackage[caption=false, font=footnotesize]{subfig}
\usepackage{epstopdf}

\usepackage{algorithm}
\usepackage{multicol}
\usepackage[noend]{algpseudocode}
\usepackage{booktabs}
\usepackage{tabularx}

\title{Monte Carlo Tree Search with Scalable Simulation Periods \\
for Continuously Running Tasks}

\author{
Seydou Ba$^1$, 
Takuya Hiraoka$^1$ $^2$,
Takashi Onishi$^1$ $^2$,
Toru Nakata$^1$, 
Yoshimasa Tsuruoka$^1$ $^3$ 
\\ 
$^1$ NEC-AIST AI Cooperative Research Laboratory\\
$^2$ NEC Security Research Laboratories\\
$^3$ The University of Tokyo\\
(ba.seydou, toru-nakata)@aist.go.jp,
t-hiraoka@ce.jp.nec.com, \\
t-onishi@bq.jp.nec.com,
tsuruoka@logos.t.u-tokyo.ac.jp
}
\begin{document}

\maketitle

\begin{abstract}
Monte Carlo Tree Search (MCTS) is particularly adapted to domains where the potential actions can be represented as a tree of sequential decisions. For an effective action selection, MCTS performs many simulations to build a reliable tree representation of the decision space. As such, a bottleneck to MCTS appears when enough simulations cannot be performed between action selections. This is particularly highlighted in continuously running tasks, for which the time available to perform simulations between actions tends to be limited due to the environment's state constantly changing.
In this paper, we present an approach that takes advantage of the anytime characteristic of MCTS to increase the simulation time when allowed. Our approach is to effectively balance the prospect of selecting an action with the time that can be spared to perform MCTS simulations before the next action selection. For that, we considered the simulation time as a decision variable to be selected alongside an action.
We extended the Hierarchical Optimistic Optimization applied to Tree (HOOT) method to adapt our approach to environments with a continuous decision space. We evaluated our approach for environments with a continuous decision space through OpenAI gym's Pendulum and Continuous Mountain Car environments and for environments with discrete action space through the arcade learning environment (ALE) platform. The evaluation results show that, with variable simulation times, the proposed approach outperforms the conventional MCTS in the evaluated continuous decision space tasks and improves the performance of MCTS in most of the ALE tasks. 
\end{abstract}

\section{Introduction}
Monte Carlo Tree Search (MCTS) \cite{coulom2006efficient,kocsis2006bandit} is a simulation-based planning method. It seeks the action that has the best expected outcome when applied to an environment at its current state.
To select an action, MCTS builds, through simulations, a search tree that represents sequences of actions that can be taken from the current state and their expected outcome.
 As such, MCTS is particularly suited to domains where actions can be represented as trees of sequential decisions, such as turn-based games and sequential decision-making tasks. Since it was proposed in 2006, MCTS has become very successful in the complex domain of computer Go \cite{gelly2007combining,silver2016mastering,silver2017mastering}. With the latest milestones of translating lessons learned from computer Go to master other board games such as Chess and Shogi \cite{silver2017masteringc}, MCTS has been solidified as an important planning method in reinforcement learning. 

Application of MCTS to many real-world environments involves selecting sequential actions for a continuous running task. In continuously running tasks, where the environment is constantly changing, MCTS is presented with a set of challenges that differ from the ones tackled in computer Go. A key challenge to MCTS in such environments is the bottleneck constituted by the number of simulations that can be performed between action selections.
The number of simulations performed between action selections dictates how accurate the search tree is in its depiction of the decision space.
 In continuously running tasks, actions are often taken to offset the effect of the constantly changing environment. As a result, the performance of MCTS is limited due to the short periods of time available for simulations between action selections.

This paper presents an approach that takes advantage of the anytime characteristic of MCTS to introduce scalable simulation periods for continuously running tasks. The proposed approach adds the simulation time as a decision variable alongside the action selection. The idea is to effectively balance the prospect of selecting an action with the time that can be spared before an action update is required. We expand the Hierarchical Optimistic Optimization applied to Tree (HOOT) method \cite{mansley2011sample} to adapt our approach to fast-changing environments with a continuous decision space. The Hierarchical Optimistic Optimization (HOO) algorithm exploits a set of promising actions that forms a general topological representation of the decision space \cite{bubeck2011x}.

The rest of the paper is organized as follows. 
First, the background to this work is reviewed.
Then, the proposed scalable search period MCTS is introduced.
Then, the evaluation performance of the proposed approach are presented.
Finally, some concluding remarks are made.

\section{Background} \label{sec:related}

\subsection{Monte Carlo Tree Search}
 The driving idea of MCTS is to determine the best action to take in the current state by representing the decision space with an incrementally growing search tree. The search tree is updated through random simulations with new simulated states (nodes) and actions (edges) iteratively added as action paths, which go from the current state to a terminal state. A node of the search tree maintains the expected value going forward from that node's state. The expected value is the average outcome of all simulations that went through the node.

MCTS simulations can be divided into multi-phase play-outs, namely, selection, expansion, roll-out, and back-propagation phases. 
Each simulation starts from the root (current state). During the selection phase, simulations go through the search tree with actions taken based on a selection policy and information maintained in the node. 
When the selection process reaches a node to which an immediate child can be added, the tree is expanded by attaching a new leaf to that node. The addition of the new node constitutes the expansion phase. Then, a roll-out policy is applied from the new leaf state to a terminal state. The straight-forward random action selection roll-out policy is widely used. Finally, the outcome of the  simulation is back-propagated to update the information maintained by the tree from the leaf node to the root. 

A key issue during the selection phase of MCTS is to balance the exploitation of promising actions and the exploration of the decision space. The commonly used Upper Confidence Bounds applied to Trees (UCT) algorithm \cite{kocsis2006bandit} offers a compromise to that. UCT is an extension of the Upper Confidence Bound (UCB) approach, which was developed for the multi-armed bandit problem \cite{auer2002finite}. 
With UCT, a node is selected to maximize the UCB1 value given as
\begin{equation} \label{eq:UCB1}
UCB1 = \bar{X_j} + C \times \sqrt{\frac{2\log n}{n_j}}
\end{equation}
where $\bar{X_j}$ is the node's value, the average of all outcomes of simulations that pass through that node. $n$ is the number of times the parent node has been visited, $n_j$ the number of times child $j$ has been visited. $C > 0$ is a coefficient, which usually
tuned experimentally to control the exploration-exploitation trade-off.

\subsection{MCTS for continuous decision spaces}
Default MCTS approaches, such as UCT, are adapted for finite-number sequential decision problems. However, as all actions from a given state are explored at least once, the look-ahead of the search tree can become very shallow when the decision space is very large. This is the case for environments with a continuous decision space where the decision space is infinite. Progressive widening and the HOOT approaches have been proposed to deal with such environments.

\subsubsection{Progressive widening}
The solution concurrently introduced as progressive widening by  Coulom \shortcite{coulom2006efficient} and progressive unpruning by Chaslot et al. \shortcite{chaslot2007progressive} initially reduces the number of evaluated actions. Eventually, more actions are added based on the number of visits to a node, and thus the decision space is progressively covered. The order of adding the actions could be determined randomly or by exploiting domain knowledge.
The progressive widening strategies assure that the added actions are sufficiently estimated, while UCT directs the tree growth toward the most promising part of the search tree.

\subsubsection{Hierarchical optimistic optimization applied to tree} 
The HOOT strategy \cite{mansley2011sample} integrates the HOO algorithm \cite{bubeck2011x} into the tree search planning to overcome the limitation of UCT in a continuous decision space. 
The HOO algorithm exploits a set of actions that forms a general topological representation of the action space as a tree.
When queried for an action, the HOO algorithm follows the path of maximal B-values, which are scores computed at the nodes. At a leaf node, an action is sampled within the range of the decision space that is represented by the leaf node.
Two child nodes are then added to the node, each covering a part of the decision space represented by the parent node.

As defined by Bubeck et al. \shortcite{bubeck2011x}, the B-value for a node $i$ is computed from its reward estimation $\hat{R_i}$ and its number of node visits $n_i$, which are saved at the node, and a reward bias based on a node's position depth $h_i$. Let $U_i$ be the upper bound on the estimate of the reward after $n$ iterations. It is given by
\begin{equation}\label{eq:Uvalue}
U_i = \hat{R_i} + \sqrt{\frac{2\log n}{n_i}} + v_{1}\rho^{h_i}
\end{equation}
for $v_1 > 0$ and $0 < \rho < 1$. 

The B-value of a node is defined as 
\begin{equation}\label{eq:Bvalue}
B_i = \min\{U_i, \max_{j \in children}\{B_j\}\},
\end{equation}
with $B_i = U_i = \infty$ for nodes that have not yet been sampled.

The HOOT approach is similar to UCT, except that HOOT places a continuous action bandit algorithm, HOO, at each node of the search tree. HOO is used to sample the decision space for action selection to overcome the discrete action limitation of UCT.

\begin{figure*}
\centering
\subfloat[\label{fig:ill_prop_mc} Mountain car example.]{\includegraphics[width=0.275\textwidth]{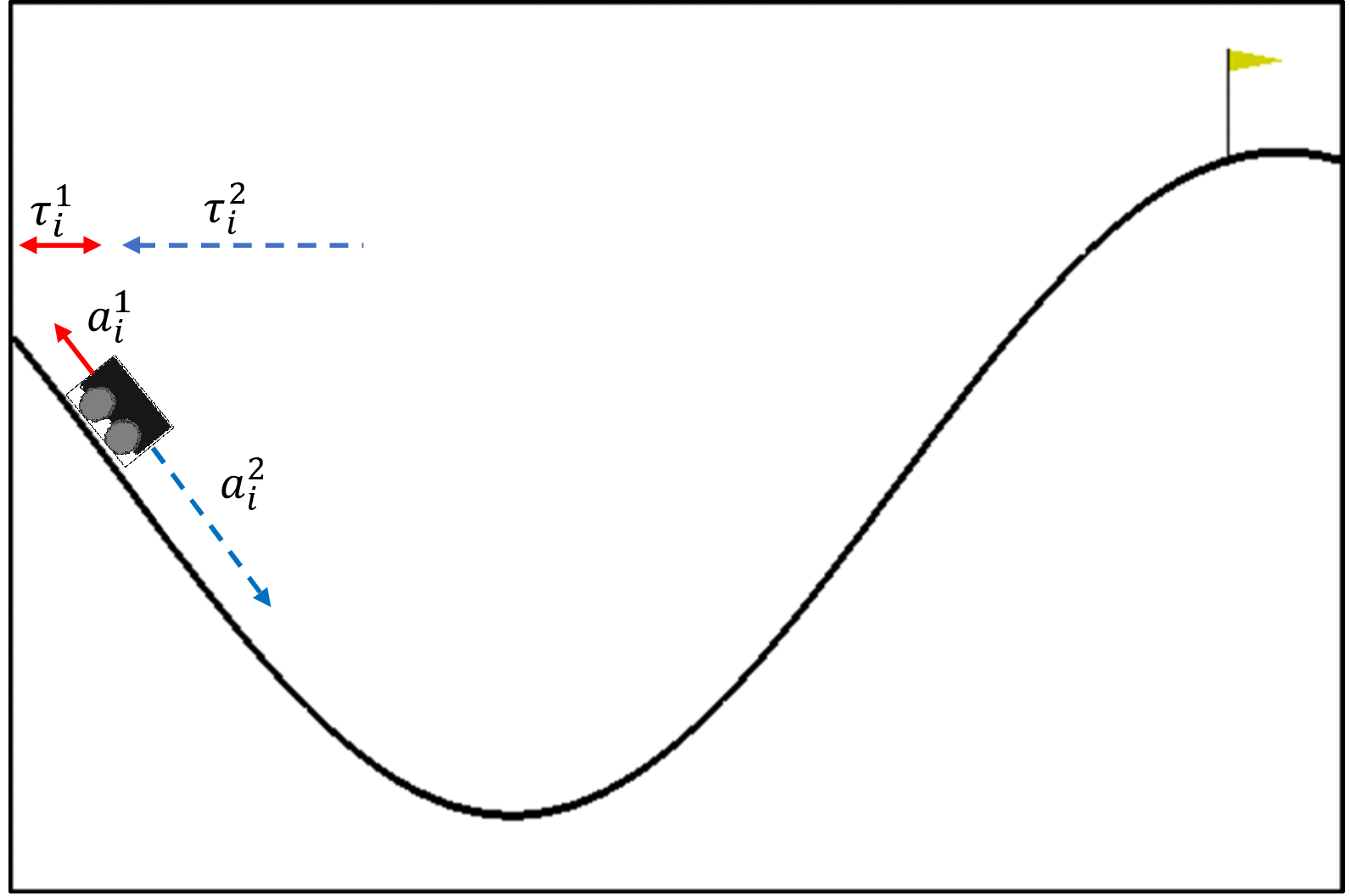}}
\hfil 
\subfloat[\label{fig:ill_prop_tl} Timeline perspective.]{\includegraphics[width=0.4\textwidth]{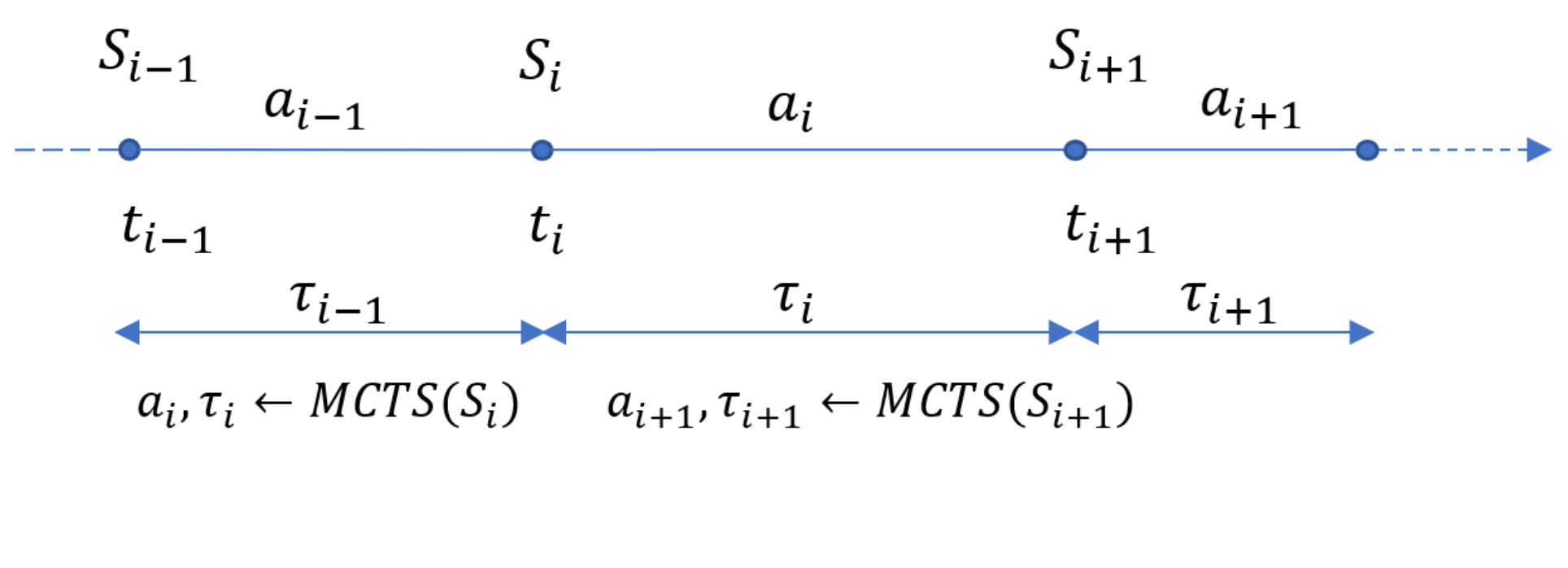}}
\hfil 
\subfloat[\label{fig:ill_prop_ts} Tree search for action and simulation period selection.]{\includegraphics[width=0.225\textwidth]{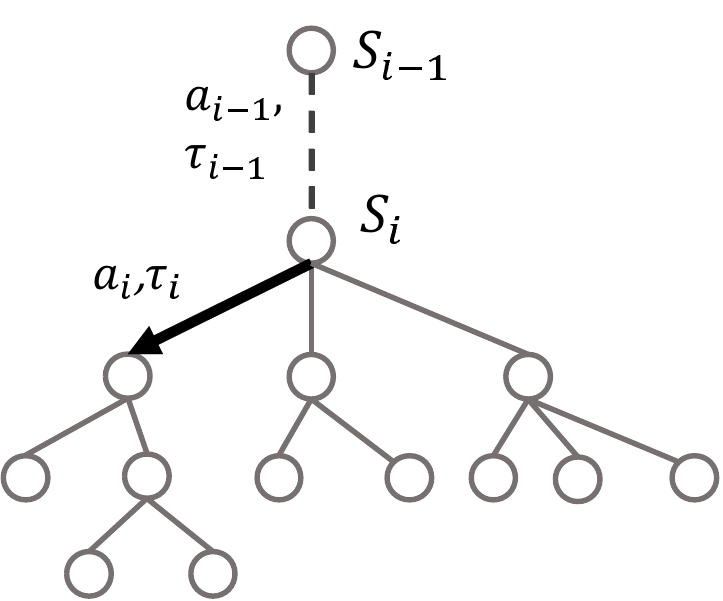}}
\caption{\label{fig:ill_prop} Scalable search period MCTS: (a) At time $t_i$, the time available for simulation is linked to the selected action, (b) While the system progresses from $S_{i-1}$ to $S_i$, MCTS simulations are performed to determine the next action and simulation period. (c) At time $t_i$, MCTS returns $(a_i,\tau_i)$ as selected action and simulation period pair from state $S_i$. 
}
\end{figure*}

\section{MCTS for continuously running tasks} \label{sec:proposed}
MCTS applications are traditionally allocated fixed time for planning between action selections. However, in continuously running environments that are constantly changing, a dilemma arises between frequently updating the action taken and allowing enough time for planning. 
With MCTS, actions are selected according to search trees that are built and updated through sampled simulations. As a search tree depicts the decision space, the efficiency of the selected action depends on the accuracy of the search tree, and therefore on the number of performed simulations. On the other hand, with limited deliberation time between actions, the number of simulations that can be performed between action selections is limited.

\subsection{Proposed scalable search approach}
To improve the performance of MCTS for continuously running tasks, we consider the possibility to run MCTS simulations with scalable deliberation periods between action selections. 
As conventional MCTS selects actions at regular steps, this confines the simulation period for all action selections to the time available between steps. We advocate to extend the time a selected action is applied before selecting a new action in order to increase the number of MCTS simulations that are performed. Our insight is that depending on the  environment's state, an action can be selected with regard to the simulation period that is allowed before an action update is necessary. 
In other words, actions can be selected while considering the period of time that would be available for simulations prior to the next action selection. 
We consider in this work the possibility for MCTS to explore the trade-off between frequent action selections and the time that can be afforded for simulations.
 
The proposed approach is to incorporate into the MCTS selection process the deliberation time between action selections as a decision variable. The allowed simulation period before the next action selection is determined alongside the selected action. 
Figure~\ref{fig:ill_prop} presents an illustration example of the proposed approach. In Figure~\ref{fig:ill_prop}\subref{fig:ill_prop_mc}, suppose that selecting either action $a_i^1$ or $a_i^2$ results in the car going up or down, respectively. In that case, the time ($\tau_i$) that can be afforded before selecting the next action depends on the selected action. If action $a_i^1$ were to be selected, a new action has to be selected as the car reaches the top of the left hill for an effective control. The proposed scalable search period MCTS take this into account to select action $a_i$ and the simulation time ($\tau_i$) that will be used for simulations to select the next action (See Fig. \ref{fig:ill_prop}\subref{fig:ill_prop_tl}-\subref{fig:ill_prop_ts}).

With the simulation time included as a variable, the transition process changes from $p(s'|s,a)$ to $p(s'|s,a,\tau)$, where $p$ is the probability of moving to state $s'$ from state $s$ if action $a$ is taken, and where $\tau$ is the period between selecting action $a$ and the next action selection. 
Note that our approach is different from a Semi-Markov Decision process where $\tau$ are random variables rather than decision variables. Note also that, with the notation $a'=(a,\tau)$, where the pair action/simulation time forms a two-dimensional action space, we revert to the default $p(s'|s,a')$.

\begin{figure*}
\centering
\includegraphics[width=0.65\textwidth]{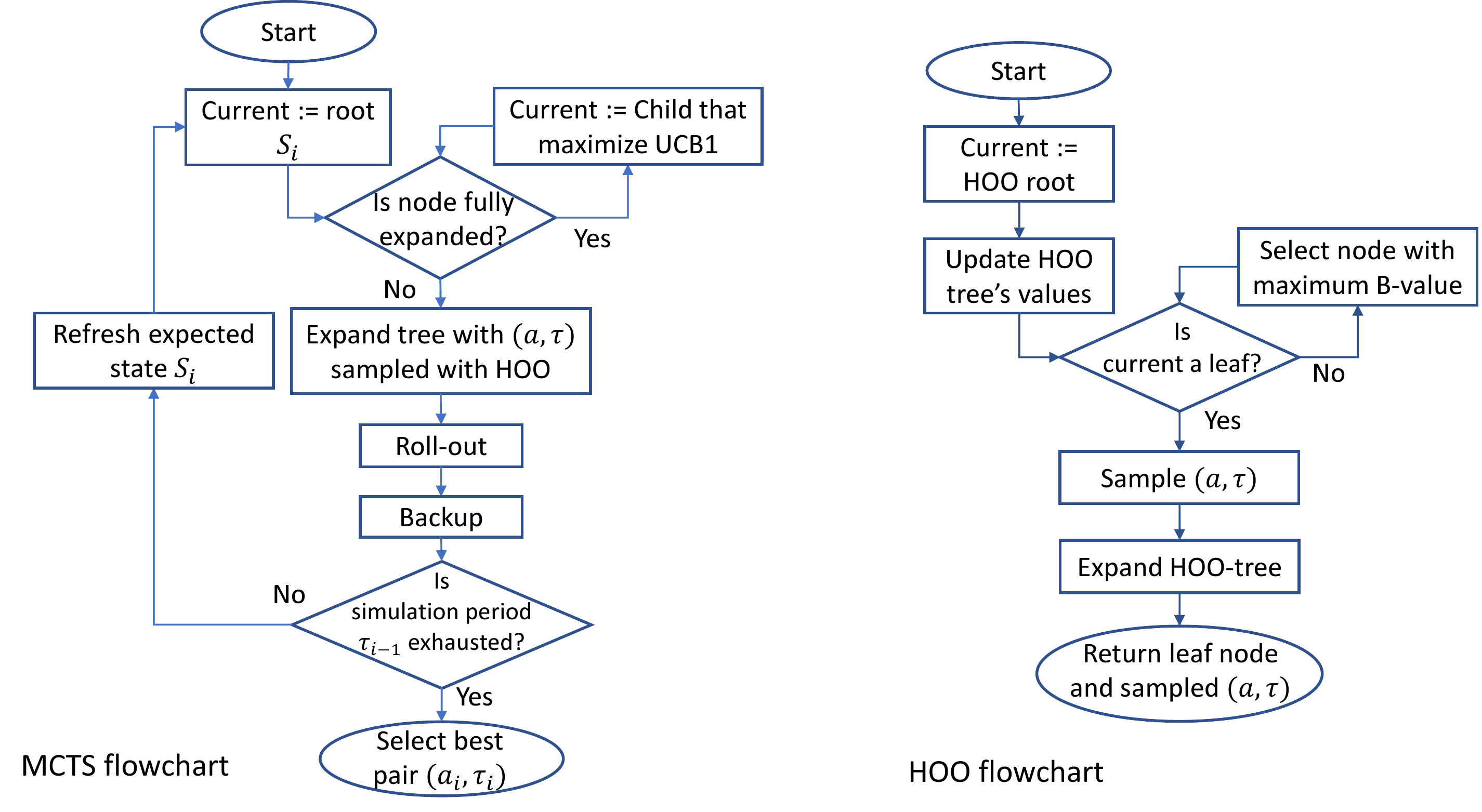}
\caption{\label{fig:chart} Algorithm flowcharts for scalable search period MCTS.
}
\end{figure*}

\subsection{Algorithm for scalable search period MCTS}
The algorithm for the proposed approach is summarily illustrated in Fig.~\ref{fig:chart}. The different phases of its implementation mostly mirror the conventional MCTS. The selection, simulation, and update phase of MCTS are mostly unchanged. The main updates are with the expansion process where the added continuous time space introduces new constraints to be tackled. 
The algorithm applies UCT with UCB1 as the selection policy to navigate the search tree. It uses progressive widening with pruning to restrict the search tree's lateral expansion. At a given time, a node is considered fully expanded if its number of children is equal to the maximum number of actions allowed with regards to the node's visits.
When an expansion is required, a HOO algorithms is queried to sample an action/deliberation time pair for the extension node.

 Progressive widening allows MCTS to initially focus the simulations on a limited number of actions to avoid having shallow search trees. As the number of performed simulations increases, it then gradually allows more actions to be considered to cover the decision space more broadly. 
We use progressive widening to decide whether a node can be further expanded.  As progressive widening limits the number of considered action/simulation time pairs considered from a node, the search tree is regularly pruned. The pruning of the search tree promotes the exploration of a vast number of action/simulation period pairs by discarding the least promising pairs in favor of trying new pairs for the search tree.

 The HOOT approach adds a filter layer to the set of action/simulation periods that are evaluated during simulations. For tasks with discrete action space, conventional HOO is used to sample the simulation periods to be paired with the actions. As for tasks with a continuous decision space, a two-dimensional HOO is introduced to sample pairs of action/simulation time. Through HOO sampling, the selection of the action/simulation period pairs to be added to the search tree are directed toward sections of the decision space where promising actions/simulation period pairs are most likely to be found. As a result, the efficiency of the actions/simulation period pairs considered in the search tree is improved, in particular when the number of simulations possible is limited.

\begin{figure*}
\centering
\subfloat[\label{fig:2d_hoo_a} Node $p$ and its HOO tree.]{\includegraphics[width=0.4\textwidth]{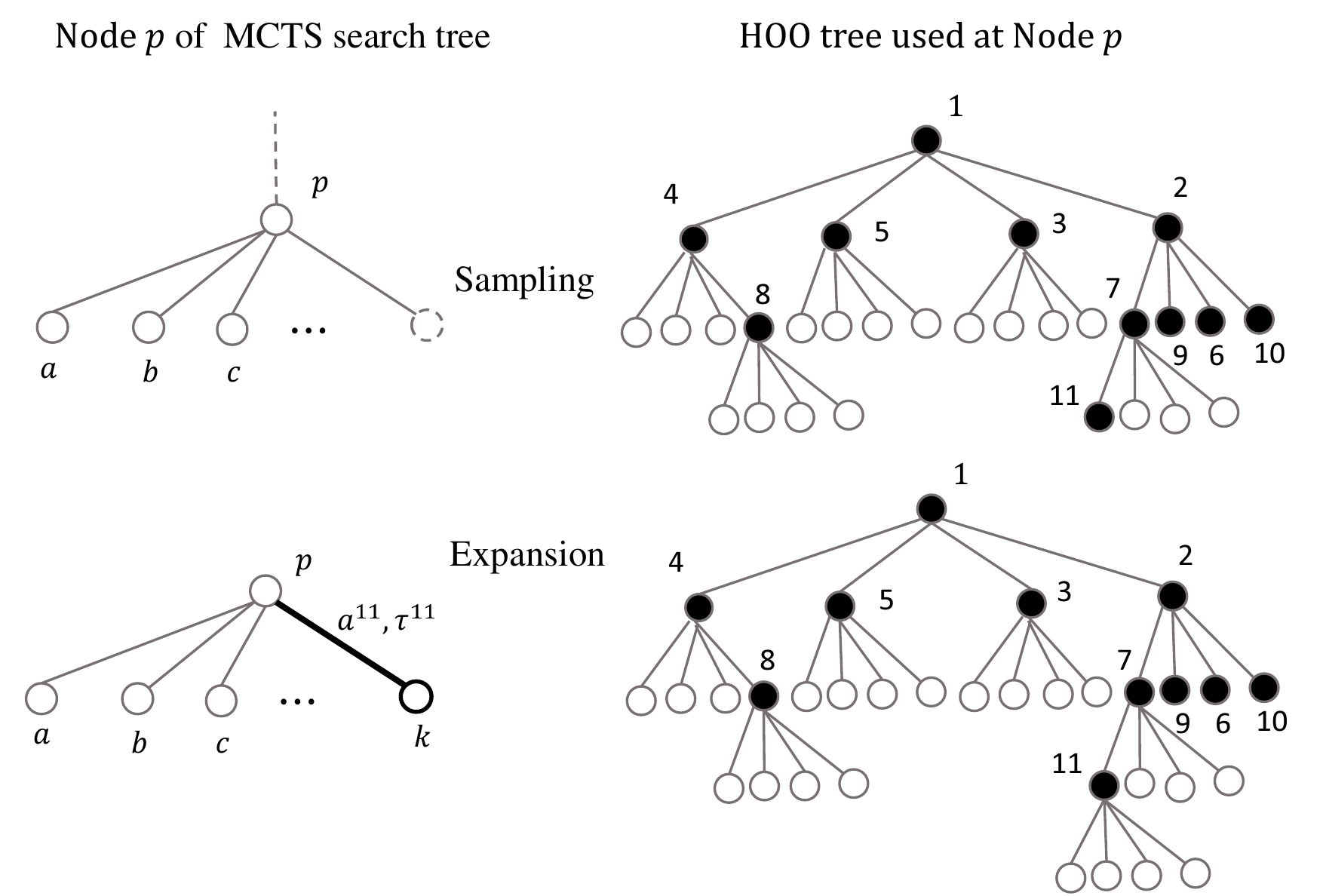}}
\hfill 
\subfloat[\label{fig:2d_hoo_b} Decision space sampled with 2-D HOO.]{\includegraphics[width=0.40\textwidth]{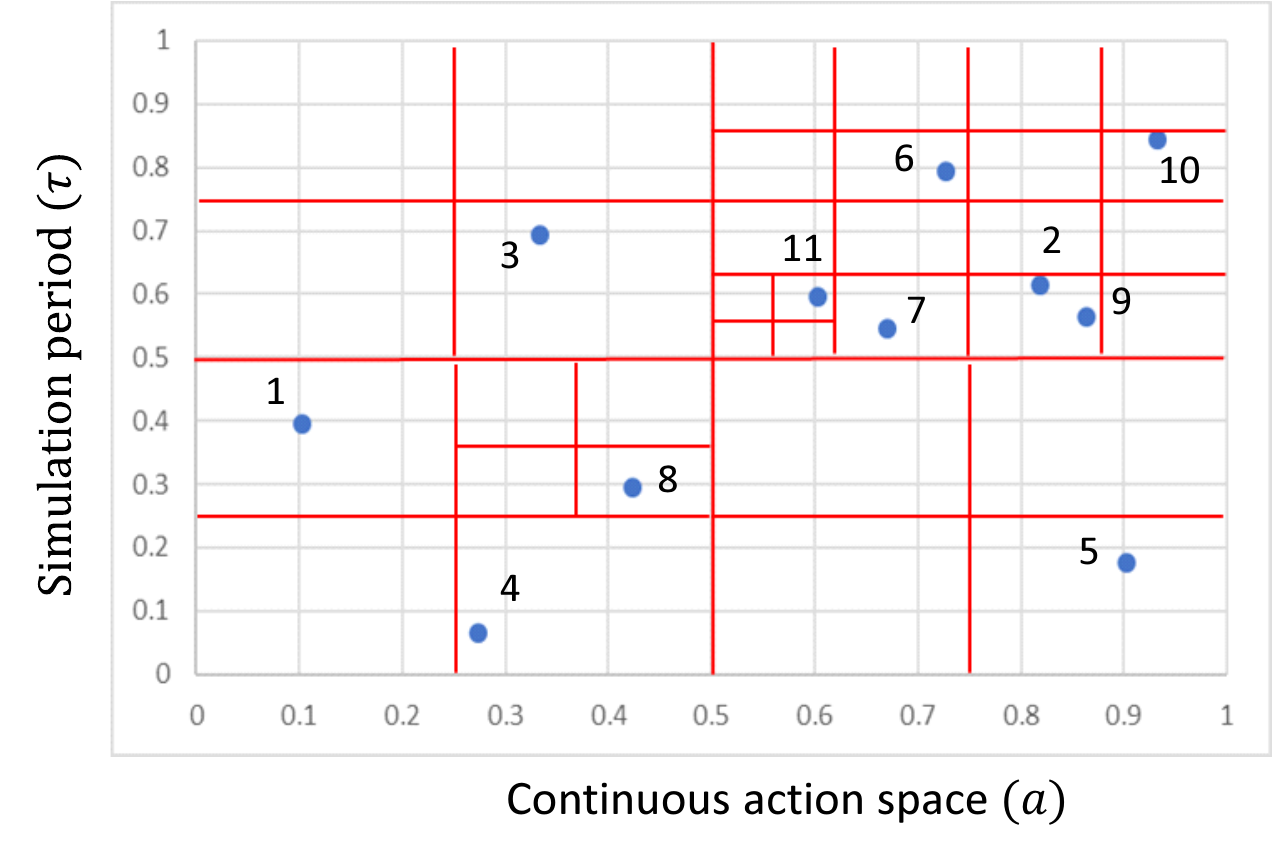}}
\caption{\label{fig:mcts_hoo} Illustration of a tree search expansion step: (a) Node $p$ of search tree is expanded with child node $k$ using the pair $(a^11,\tau^11)$ sampled by querying the HOO algorithm. (b) 2-D representation of the HOO tree. Each node of the HOO tree covers a given region, and each of its children cover a sub-area of its region.}
\end{figure*}

Another aspect considered in the proposed approach for  continuously running tasks is the effect of running MCTS simulations for an expected state that is multiple steps ahead. In continuously running tasks, the MCTS simulations to select an action/simulation period pair ($a_i, \tau_i$) are performed considering the state $S_i$ that is expected after applying the previously selected action $a_{i-1}$ for the simulation period $\tau_{i-1}$ (See Fig. \ref{fig:ill_prop_tl}).
 When the simulation period exceeds a single step, the expected state at the end of the simulations may drift from the task's actual state. This occurs if the task is stochastic or if the simulation model for a deterministic task is not $100\%$ accurate.
To mitigate this effect on the accuracy of the search tree, the task's state during the MCTS simulations is monitored and the expected state for the simulation model is updated after each step.

\subsection{2-D HOOT for scalable search period MCTS}
We adapt the HOOT method to our approach to set scalable search periods for MCTS. For tasks with a discrete action space, HOOT is applied as defined by Bubeck et al. \shortcite{bubeck2011x} to sample the simulation periods to be paired with the actions. However, for tasks with a continuous decision space, both actions and simulation periods are to be sampled. For that, the HOO algorithm is updated to work in a two-dimensional space (See Fig.~\ref{fig:mcts_hoo}). Mainly, the HOO algorithm is structurally changed as the HOO tree can no longer be handled as a binary tree. Bifurcating both the decision space and the time space covered by a node leads to four regions being created, with each region represented by an additional child node.

 In our implementation of the approach, each node of the MCTS contains a HOO tree from which the action/deliberation time pairs $(a,\tau)$ used as transition edges to its child nodes are sampled. When queried during an expansion of the search tree, the HOO algorithm follows the path of largest B-values to a leaf node where it samples a pair of action/deliberation time from the node's covered range ($a \in [a_{min}, a_{max}], \tau \in [\tau_{min}, \tau_{max}]$).
 
 Each node that is added to the search tree saves a pointer to its parent node's HOO tree, parallel with initializing a new HOO tree.
 The pointers to the parent nodes' HOO trees are saved for updating the HOO tree's average values. The value of a HOO node is taken as the  average reward pulled from the tree search instead of the immediate reward returned by directly applying that action. In the case of HOOT, the immediate reward does not provide information about the effectiveness of the selected action with regard to the MCTS. Using the average expected value leads the HOO selection toward a section of the decision space where actions are expected to perform well in the long run.

In the original HOOT paper \cite{mansley2011sample}, the action was taken by greedily following branches according to the mean rewards as opposed to the B-values. In this paper, we compute the B-value from the average values given by the MCTS. 
Since the MCTS values change as the simulations progress, the HOO values are updated to mirror them. Whenever the value of a tree search node is updated, the value of its corresponding HOO node is also updated.
The value from the MCTS corresponds to the value of the action selected from that HOO node. However, the value of a HOO node reflects on all actions selected from nodes of its sub-tree. As such, the values $\hat{R_i}$ of the HOO nodes are iteratively updated to reflect that. They are given by 
\begin{equation}
\hat{R_i} = \frac{\bar{X_i} + \sum_{j \in children} {n_j \times \hat{R}_j}}{n_i}
\end{equation}
where $\bar{X_i}$ is the action value from the MCTS tree, and $n_i$ is the number of visits.

The $U$ and $B$ values are computed as given by Equations~(\ref{eq:Uvalue})~and~ (\ref{eq:Bvalue}).
To avoid unnecessary repetitions of the iterative updates, their updates are performed only when an action selection is required.  

\section{Performance evaluation} \label{sec:eval}   
We evaluated the performance of the proposed scalable search period MCTS (SSP-MCTS) for tasks with a continuous decision space through OpenAI gym's Pendulum and Continuous Mountain Car environments~\cite{brockman2016openai} and for environments with a discrete action space through the arcade learning environment (ALE) platform~\cite{bellemare2013arcade}.  The experiments were performed on servers consisting of Intel cores Xeon E5-2690v4, with $2.6$ Ghz clock speed. 

\subsection{Classical control with continuous decision space} \label{sec:eval_cc}
We assess the performance of the proposed approach in terms of average accumulated rewards per episode. Our results are compared to the performance of conventional MCTS. For that, we considered two cases of conventional MCTS implementations where the deliberation time is not a factor. In the first case only progressive widening (\emph{PW}) is used,  and  in the second case progressive widening is paired with conventional HOOT (\emph{PW+HOOT}). 

\begin{figure*}[t]
\centering
\subfloat[\label{fig:rslt_pend} Average accumulated rewards in Pendulum.]{\includegraphics[width=0.325\textwidth]{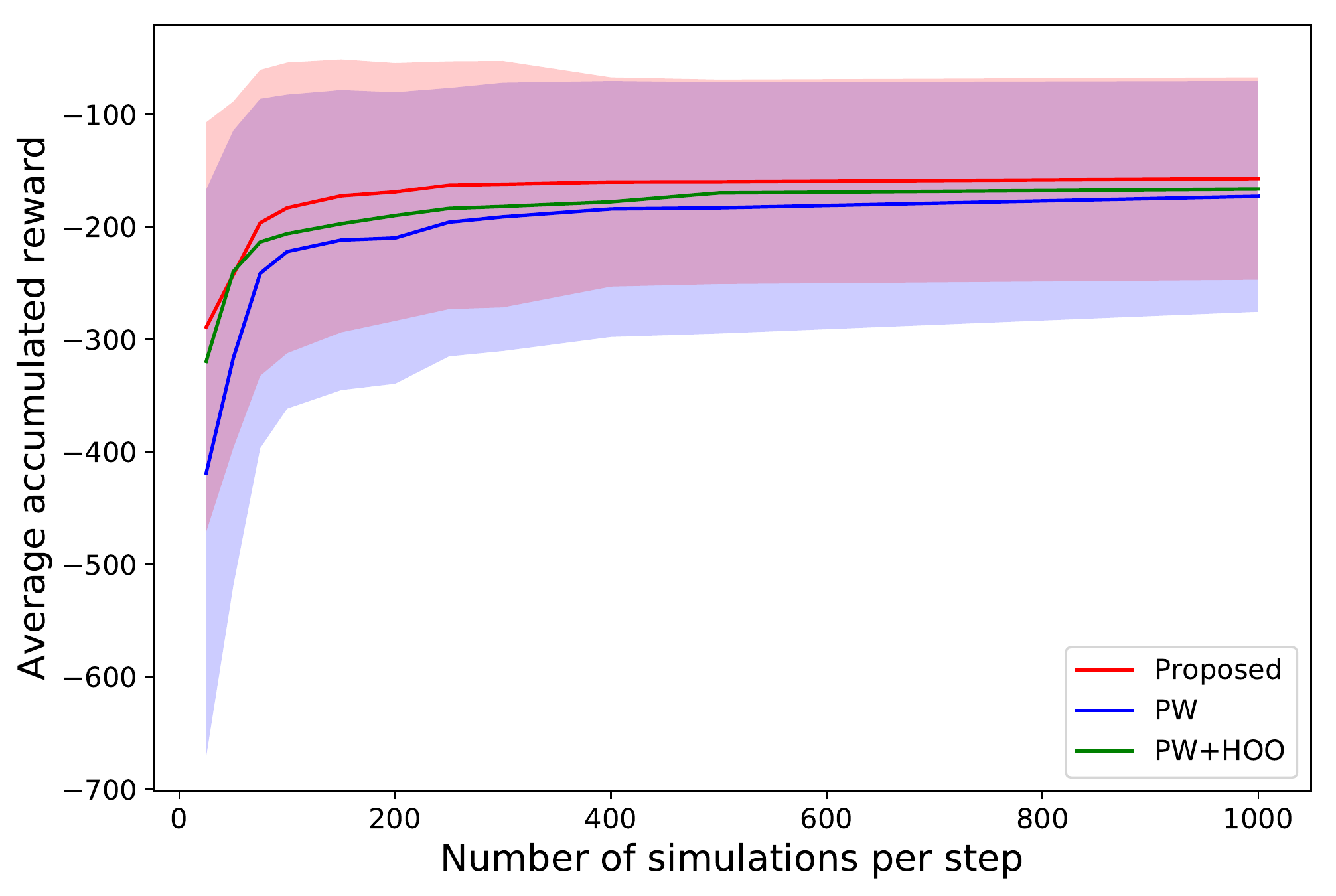}}
\hfill
\subfloat[\label{fig:rslt_cmc} Average accumulated rewards in Continuous mountain car.]{\includegraphics[width=0.325\textwidth]{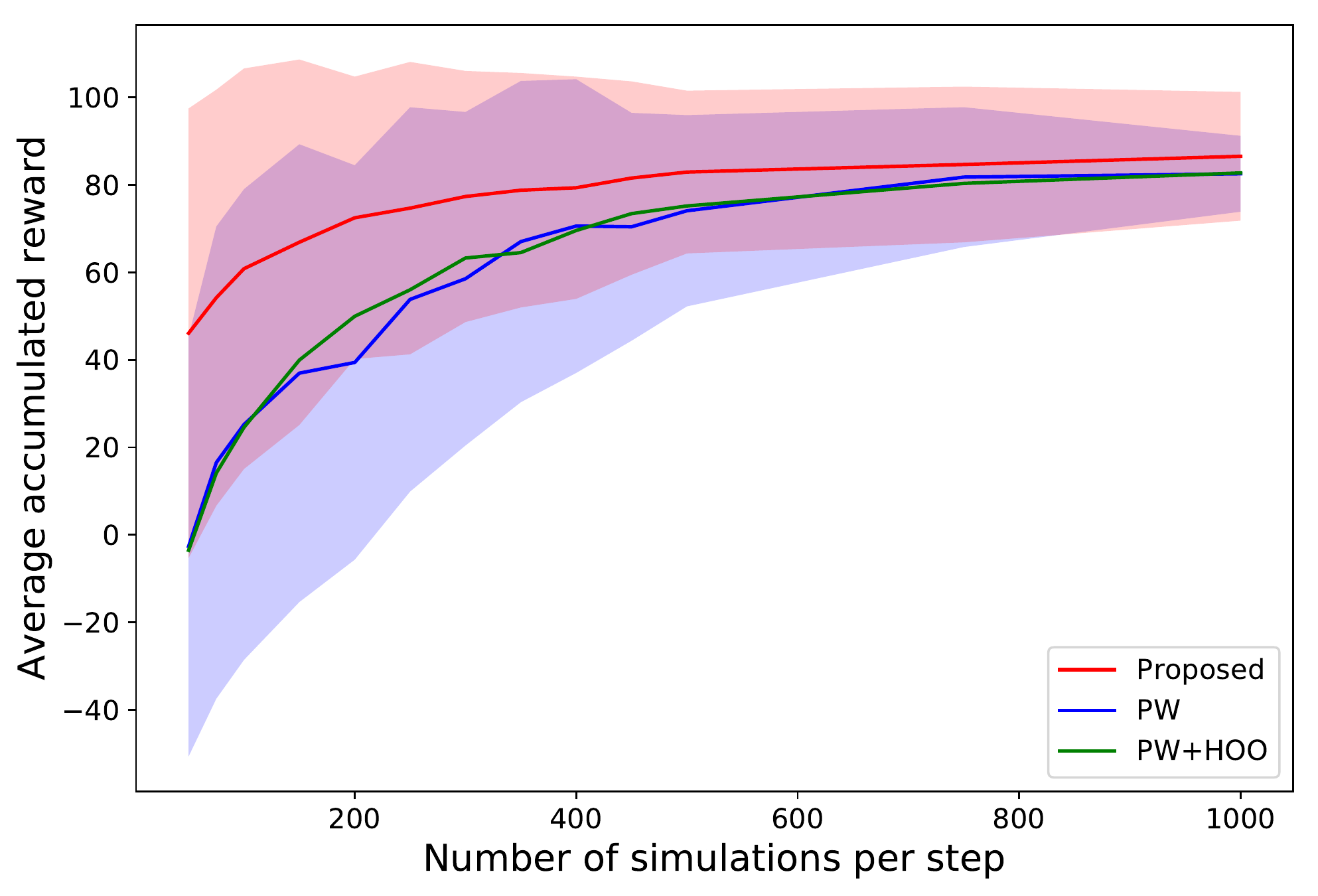}}
\hfill
\begin{minipage}[b]{.275\textwidth}
\centering
\subfloat[\label{fig:tau_pend} Distribution of selected simulation period $\tau$.]{\includegraphics[width=0.975\textwidth]{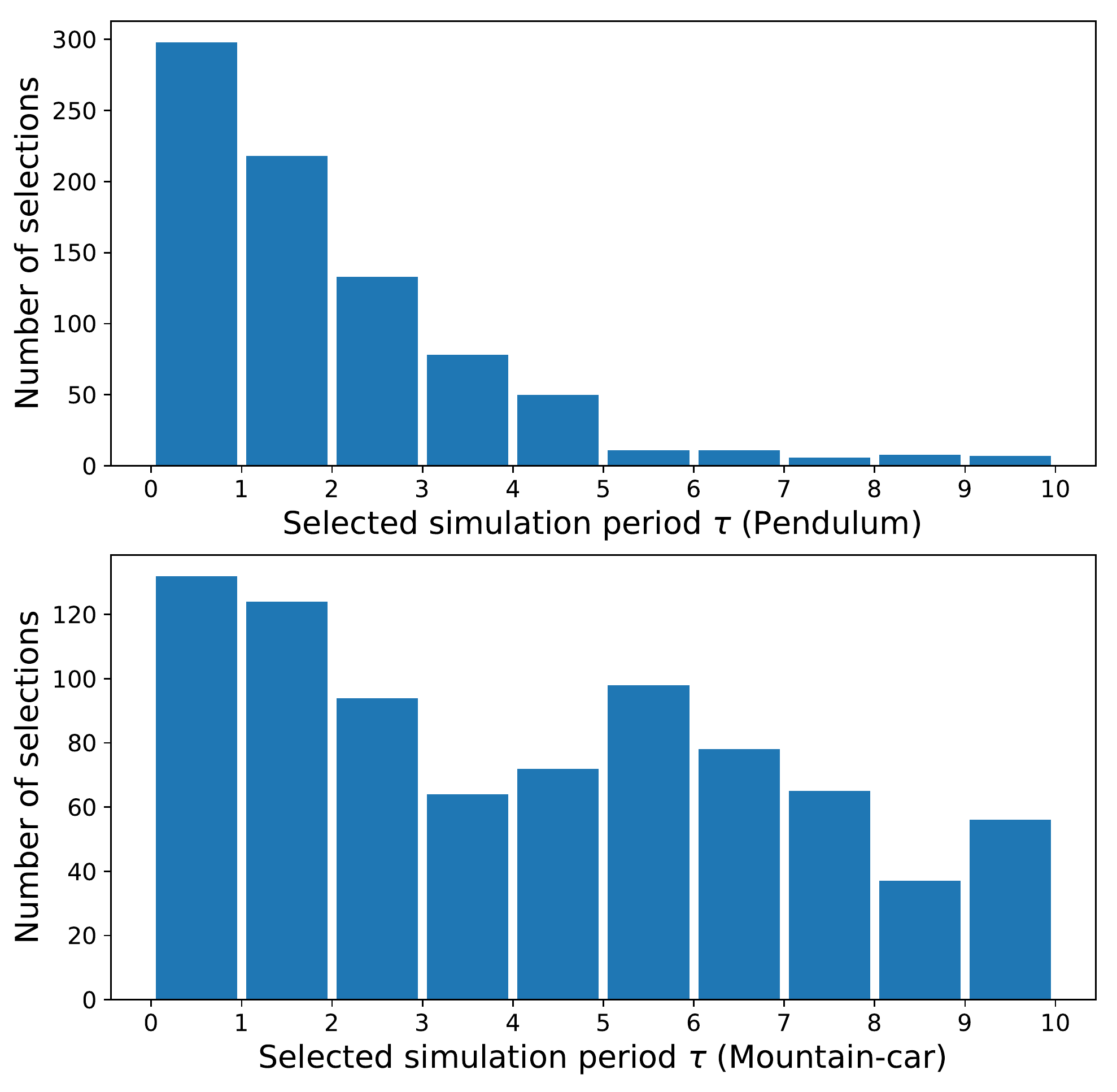}}
\end{minipage}
\caption{\label{fig:rslt_rew} Evaluation results for tasks with a continuous action space.}
\end{figure*}

Two OpenAI gym~\cite{brockman2016openai} environments, Pendulum and Continuous Mountain Car environments, are used for our simulation purpose. 
With the Pendulum environment, the goal is to keep a frictionless pendulum standing up. The pendulum starts in a random position, and continuous-value torques must be selected to swing it up so it stays upright.
With the mountain car environment, a car is on a one-dimensional track, positioned between two hills. The goal is to drive up the hill on the right; however, the car is under-powered. Therefore, to build up momentum and accelerate towards the target, the opposite hill must be climbed.

We set the number of simulations that can be performed per unit of time as a simulation parameter rather than the run time. 
Given a uniform processing speed, the number of performed simulations is proportional to the run time. Furthermore, the performance represented as a function of the number of simulations per time step is not affected by fluctuations of the processing speed due to events unrelated to the simulation (e.g. computer load).

The proposed SSP-MCTS approach outperforms conventional MCTS in both considered environments as illustrated in Figure~\ref{fig:rslt_rew}. Figure~\ref{fig:rslt_rew} presents the average accumulated rewards per episode for the different MCTS approaches with regard to the number of simulations per step. The number of simulation episodes is $5000$ for the pendulum environment and $500$ for the continuous mountain car environment. 
We observe from Figure~\ref{fig:rslt_rew}\subref{fig:rslt_pend} the effect of default HOOT on conventional MCTS. Adding to the progressive widening method the HOOT approach, which prioritizes the evaluation of the most promising actions, improves the performance of conventional MCTS. Our approach achieves further improvement by including the updated HOOT method to flexibly select simulation periods during action selection. 

The advantage of the SSP-MCTS over conventional MCTS is more visible in Figure~\ref{fig:rslt_rew}\subref{fig:rslt_cmc}. This suggests that the proposed approach is more efficient with the continuous mountain car environment than it is with the pendulum environment. However, Figure~\ref{fig:rslt_rew}\subref{fig:tau_pend} suggests that this result is due to intrinsic characteristics of the environments rather than to a lack of efficiency from the proposed approach. In fact, Figure~\ref{fig:rslt_rew}\subref{fig:tau_pend} demonstrates the versatility of the proposed approach. 

Figure~\ref{fig:rslt_rew}\subref{fig:tau_pend} presents the distribution of the simulation periods selected by the SSP-MCTS approach. We observe that our approach was able to identify when applying action updates in short succession was preferable and when it was more advantageous to run extended simulations before updating the action. The Pendulum environment is a stability task where   frequent updates are required to keep a steady control of the pole. That is not the case of the continuous mountain car environment, where the car is constantly moving from one side of the hill to the other until it reaches the goal. 
The two environments offer different types of challenge which are reflected in the distribution of the selected simulation periods. Short simulation periods are selected for the pendulum environment to ensure continuous control, while the selected simulation periods are quite distributed for the continuous mountain car.

\subsection{ALE for discrete action space evaluation} \label{sec:eval_da}
The performance of the proposed SSP-MCTS for tasks with a discrete action space are evaluated using the ALE environment with $54$ Atari $2600$ games. The results are presented in comparison with UCT and IW(1). The Iterated Width (IW) algorithm has been introduced as a classical planning algorithm that takes a planning problem as an input, and computes an action sequence that solves the problem as the output  \cite{geffner2012width}. Its variant  algorithms, IW(1) and 2BFS, have been implemented for Atari games and reported in Lipovetzky, Ramirez, and Geffner~\shortcite{lipovetzky2015classical}.

To follow their experimental setup, we limit the maximum number of simulated frames per step to $150000$. The number of MCTS simulations per step is $500$ and the maximum search depth is $300$ frames. The discount factor used is $\gamma = 0.99$. Due to the wide range of scores throughout the different games, the exploration constant $C$ is selected among $0.1$, $0.01$ and $0.001$ depending on the game.
For the proposed approach, each presented result is the rounded average performance over at least $10$ episodes run.

Table~\ref{tab:sample} presents some preliminary results of the proposed SSP-MCTS in comparison with UCT in selected games. Both approaches were simulated under our experimental settings, with the maximum search depth limited to a $100$ frames. 
These results indicate that our experimental settings lead to results that are similar to previously reported ones \cite{bellemare2013arcade}. 
As such, for the whole set of games, we compare the scores obtained for the proposed approach with the ones for UCT and IW(1) as reported in Bellemare et al.~\shortcite{bellemare2013arcade} and~Lipovetzky, Ramirez, and in Geffner~\shortcite{lipovetzky2015classical}, respectively  (see Table~\ref{tab:all}). IW(1) is selected out of the iterative width algorithms as it has overall the best reported performance among them.

\begin{table}
  \centering
  \begin{small}
    \begin{tabular}{|l|r|r|r|r|}
\hline
&  \multicolumn{2}{c|}{SSP-MCTS} & \multicolumn{2}{c|}{UCT}\\ \hline
\textbf{Game} & Score & Time (s) & Score & Time (s) \\ \hline
Asterix & 81950 & 33 & \textbf{564000} & 30 \\ \hline
Beamrider & \textbf{4830} & 44 & 3945 & 35 \\ \hline
Freeway & \textbf{3} & 61 & 0 & 55 \\ \hline
Seaquest & \textbf{915} & 41 & 734 & 38 \\ \hline
SpaceInvaders & \textbf{2697} & 37 & 2475 & 33 \\ \hline
\end{tabular}
  \end{small}
    \caption{Scores comparison with UCT in selected games with the average simulation time per step}
    \label{tab:sample}
\end{table}

\begin{table}
  \centering
  \begin{small}
    \begin{tabular}{|l|r|r|r|}
\hline
Game & SSP-MCTS & UCT & IW(1) \\ \hline
Alien & 8823 & 7785 & \textbf{25634} \\ \hline
Amidar & \textbf{3224} & 180 & 1377 \\ \hline
Assault & \textbf{3430} & 1512 & 953 \\ \hline
Asterix & 120740 & \textbf{290700} & 153400 \\ \hline
Asteroids & 5974 & 4661 & \textbf{51338} \\ \hline
Atlantis & \textbf{304550} & 193858 & 159420 \\ \hline
BankHeist & 525 & 498 & \textbf{717} \\ \hline
BattleZone & \textbf{590750} & 70333 & 1160 \\ \hline
BeamRider & 8951 & 6625 & \textbf{9108} \\ \hline
Berzerk & 704 & 554 & \textbf{2096} \\ \hline
Bowling & \textbf{110} & 25 & 69 \\ \hline
Boxing & 99 & 100 & 100 \\ \hline
Breakout & 338 & 364 & \textbf{384} \\ \hline
Carnival & 6034 & 5132 & \textbf{6372} \\ \hline
Centipede & \textbf{172544} & 110422 & 99207 \\ \hline
ChopperCommand & \textbf{38060} & 34019 & 10980 \\ \hline
CrazyClimber & \textbf{123500} & 98172 & 36160 \\ \hline
DemonAttack & \textbf{33434} & 28159 & 20116 \\ \hline
DoubleDunk & 24 & 24 & -14 \\ \hline
ElevatorAction & \textbf{23264} & 18100 & 13480 \\ \hline
Enduro & 358 & 286 & \textbf{500} \\ \hline
FishingDerby & \textbf{49} & 38 & 30 \\ \hline
Freeway & 11 & 0 & \textbf{31} \\ \hline
Frostbite & \textbf{1912} & 271 & 902 \\ \hline
Gopher & 13846 & \textbf{20560} & 18256 \\ \hline
Gravitar & \textbf{5500} & 2850 & 3920 \\ \hline
Hero & \textbf{15061} & 12860 & 12985 \\ \hline
IceHockey & 46 & 39 & \textbf{55} \\ \hline
JamesBond & 2527 & 330 & \textbf{23070} \\ \hline
JourneyEscape & 14400 & 7683 & \textbf{40080} \\ \hline
Kangaroo & 1850 & 1990 & \textbf{8760} \\ \hline
Krull & 5995 & 5037 & \textbf{6030} \\ \hline
KungFuMaster & 44600 & 48855 & \textbf{63780} \\ \hline
MontezumaRevenge & 0 & 0 & 0 \\ \hline
MsPacman & 16702 & \textbf{22336} & 21695 \\ \hline
NameThisGame & \textbf{29770} & 15410 & 9354 \\ \hline
Pong & 17 & 21 & 21 \\ \hline
Pooyan & \textbf{28625} & 17763 & 11225 \\ \hline
PrivateEye & \textbf{965} & 100 & -99 \\ \hline
Q*Bert & 16485 & \textbf{17343} & 3705 \\ \hline
Riverraid & 4479 & 4449 & \textbf{5694} \\ \hline
RoadRunner & 30175 & \textbf{38725} & 94940 \\ \hline
RobotTank & \textbf{78} & 50 & 68 \\ \hline
Seaquest & 1493 & 5132 & \textbf{14272} \\ \hline
SpaceInvaders & \textbf{4420} & 2718 & 2877 \\ \hline
StarGunner & \textbf{7630} & 1207 & 1540 \\ \hline
Tennis & 2 & 3 & \textbf{24} \\ \hline
TimePilot & \textbf{64325} & 63855 & 35000 \\ \hline
Tutankham & \textbf{246} & 226 & 172 \\ \hline
UpAndDown & 105603 & 74474 & \textbf{110036} \\ \hline
Venture & 0 & 0 & \textbf{1200} \\ \hline
VideoPinball & \textbf{854894} & 254748 & 388712 \\ \hline
WizardOfWor & \textbf{126500} & 105500 & 121060 \\ \hline
Zaxxon & \textbf{52800} & 22610 & 29240 \\ \hline
\hline
Times Best  & 25 & 6 & 20 \\ \hline
Better than UCT & 40 & - & 31 \\ \hline
\end{tabular}
  \end{small}
    \caption{Scores of proposed SSP-MCTS in comparison with UCT and IW(1).}
    \label{tab:all}
\end{table}

Table~\ref{tab:all} presents the score of the proposed SSP-MCTS in comparison with UCT and IW(1). The SSP-MCTS performs better than both UCT and IW(1) on $25$ of $54$ games ($6$ for UCT, $20$ for IW($1$)). More interestingly, SSP-MCTS outperforms UCT on $40$ games ($31$ for IW(1)). In most cases, the SSP-MCTS performs significantly better. The improvements over UCT scores, have led the SSP-MCTS to outscore IW($1$) on ten games for which UCT had lower scores than IW($1$) (e.g. Bowling, Frostbite, VideoPinball). 
These results indicates that the extended simulation periods allow the proposed approach to select more effective actions. 
We note that repeating the same action during the extended simulation periods offer some advantage in games where the player benefits from repeating a particular action, e.g. accelerating in Enduro, or moving up in Freeway. That being said, the wide range of games where SSP-MCTS outperforms UCT, and in which there is no apparent advantage in repeating the same action (e.g. Bowling, Demon Attack, Fishing Derby, Private Eye, Space Invader), consolidates the proposed approach.

On the other hand, there are ten games where the SSP-MCTS have scored lower than UCT. In five of those ten games, namely Asterix, gopher, pacman, road runner and seaquest, a sizable drop of score is noticed. These are games where a failure to select the right action in some states will lead to the loss of a life for the player. Since the losses of lives are not accounted for until the last one, the loss of which ends the game, repeating actions by the proposed SSP-MCTS sometimes leads to early game termination. Theoretically, the SSP-MCTS could select one-step simulation periods to at least match the score of UCT. However, with the simulation time added as a decision variable and progressive widening used to limit the number of considered action/simulation time pairs, there is a probability of missing out as the cost of exploring different simulation time for multiple actions.

\section{Related work}
MCTS has proven beneficial in a wide range of domains. An extensive survey of early applications of MCTS is given by Browne et al. \shortcite{browne2012survey}. 
MCTS is used for real-time game environments to control the Pac-Man
character \cite{pepels2014real,guo2014deep} and as an offline planner in an approach that combines it with DQN in the ALE \cite{guo2014deep}.
Silver and Veness \shortcite{silver2010monte} introduced a Monte-Carlo algorithm for online planning in large partially observable Markov decision problems (POMDPs) and their method was extended to Bayes-Adaptive POMDPs by Katt et al. \shortcite{katt2017learning}.
MCTS was also applied for stochastic environments \cite{couetoux2013monte,yee2016monte}.
Couetoux \shortcite{couetoux2013monte} advocated the use of double progressive widening for stochastic and continuous sequential decision making problems.
Yee et al. \shortcite{yee2016monte} proposed a variant of MCTS based on Kernel Regression KR-UCT for continuous action spaces with execution
uncertainty.
They based their approach on the existence of similarities among actions that could generate a common outcome.

Dynamic Frame skip Deep Q-Network (DFDQN) \cite{lakshminarayanan2016dynamic} has been considered for the ALE environment. 
DFDQN treats the frame skip rate as a dynamic learnable parameter that defines the number of times a selected action is repeated based on the current state. The agent can select a pair of action/frame skip rate from a set of options that includes two predefined frame skip rate values for each action.
 In comparison, our approach proposes scalable simulation time, during which a selected action is repeated, for MCTS. The simulation periods are not predefined but selected alongside the actions through MCTS simulations.

\section{Conclusions}   \label{sec:conclusions}
This paper has proposed a scalable search period MCTS approach that balances action selections with the simulation time that can be afforded for effective action selections in continuously running tasks. To mitigate the trade-off between action selection frequency and the time available for MCTS simulations, the proposed approach considers the simulation time available between action selections as a decision variable to be selected alongside the actions.
To direct the MCTS towards the most promising area of the decision space, the implementation algorithm relies on progressive widening, pruning and HOOT. An updated HOOT  is introduced for action/simulation time pairs sampling in tasks with a continuous decision space. 
The simulation results suggest that the proposed scalable search period MCTS approach effectively selects actions/simulation time pairs with regard to the environment. The MCTS with scalable simulation periods outperforms the conventional MCTS in simulated continuous action space environments and improve its result in most of the Atari games.

\newpage

\bibliographystyle{aaai}
\bibliography{aaai19}

\end{document}